\documentclass[10pt,twocolumn,letterpaper]{article}

\usepackage[pagenumbers]{cvpr}  

\usepackage{graphicx}
\usepackage{caption}
\usepackage{subcaption}
\usepackage{float}
\usepackage{longtable}
\usepackage{makecell}
\usepackage{listings}
\usepackage{adjustbox}
\usepackage{xcolor}
\usepackage{amsmath}

\usepackage{tikz}
\usetikzlibrary{fit, positioning}
\usepackage{multirow} 
\usetikzlibrary{shapes.geometric, arrows, positioning}
\tikzstyle{block} = [rectangle, draw, rounded corners, text centered, text width=4cm, minimum height=1.5cm]
\tikzstyle{dataset} = [block, fill=green!30, minimum height=1cm]
\tikzstyle{augment} = [block, fill=blue!30, minimum height=1cm]
\tikzstyle{model} = [block, fill=yellow!30, text width=3.5cm, minimum height=1cm]
\tikzstyle{evaluation} = [block, fill=red!30, minimum height=1cm]
\tikzstyle{arrow} = [thick,->,>=latex] 

\definecolor{cvprblue}{rgb}{0.21,0.49,0.74}
\usepackage[pagebackref,breaklinks,colorlinks,allcolors=cvprblue]{hyperref}

\title{AugmentGest: Can Random Data Cropping Augmentation Boost Gesture Recognition Performance?}

\author{Nada Aboudeshish \hspace{2.4cm} Dmitry Ignatov \hspace{2.4cm} Radu Timofte \\\\
Computer Vision Lab, CAIDAS, University of Würzburg, Germany}

\begin{document}
\maketitle

\begin{abstract}
Data augmentation is a crucial technique in deep learning, particularly for tasks with limited dataset diversity, such as skeleton-based datasets. This paper proposes a comprehensive data augmentation framework that integrates geometric transformations, random cropping, rotation, zooming and intensity-based transformations, brightness and contrast adjustments to simulate real-world variations. Random cropping ensures the preservation of spatio-temporal integrity while addressing challenges such as viewpoint bias and occlusions. The augmentation pipeline generates three augmented versions for each sample in addition to the data set sample, thus quadrupling the data set size and enriching the diversity of gesture representations. 
The proposed augmentation strategy is evaluated on three models: multi-stream e2eET, FPPR point cloud-based hand gesture recognition (HGR), and DD-Network. Experiments are conducted on benchmark datasets including DHG14/28, SHREC’17, and JHMDB. The e2eET model, recognized as the state-of-the-art for hand gesture recognition on DHG14/28 and SHREC’17. The FPPR-PCD model, the second-best performing model on SHREC’17, excels in point cloud-based gesture recognition. DD-Net, a lightweight and efficient architecture for skeleton-based action recognition, is evaluated on SHREC’17 and the Human Motion Data Base (JHMDB).
The results underline the effectiveness and versatility of the proposed augmentation strategy, significantly improving model generalization and robustness across diverse datasets and architectures. This framework not only establishes state-of-the-art results on all three evaluated models but also offers a scalable solution to advance HGR and action recognition applications in real-world scenarios. The framework is available at \href{https://github.com/NadaAbodeshish/Random-Cropping-augmentation-HGR}{https://github.com/NadaAbodeshish/Random-Cropping-augmentation-HGR}.

\end{abstract}

\section{Introduction}
Gesture recognition has become a cornerstone of human-computer interaction, unlocking new possibilities in applications such as gaming, virtual and augmented reality, robotics, and assistive technologies. As an intuitive method of communication, gesture recognition plays a pivotal role in enhancing user experiences and enabling natural interactions with technology. Despite significant advancements in deep learning and computer vision, developing robust and generalisable gesture recognition systems remains challenging. Factors such as limited dataset diversity, variations in viewpoints, and real-world complexities continue to hinder the performance of gesture recognition models.
Data augmentation has been shown to be a powerful technique for overcoming these challenges by artificially increasing the variability and size of training datasets, thus enhancing model robustness and generalisation. Techniques such as geometric transformations and intensity-based adjustments have demonstrated significant promise in image-based tasks. However, their potential for improving skeleton-based gesture recognition, where maintaining spatio-temporal and structural integrity is crucial, has not been fully explored. Augmentation strategies that preserve the dynamic and sequential nature of skeletal data are critical to achieving reliable recognition performance in gesture-based applications.
This paper propose a data augmentation framework Tailored for skeleton-based gesture recognition. The proposed framework combines geometric transformations, random cropping, and intensity-based modifications to address the diversity of the data set and the variability of the real world. The augmented data sets are evaluated with a state-of-the-art model and two highly competitive models—\textbf{e2eET}, \textbf{FPPR-PCD}, and \textbf{DD-Net} \cite{10216066,bigalke2021fusing,yang2019ddnet}—using benchmark data sets, DHG14/28, SHREC17, and JHMDB \cite{DeSmedt2016,DeSmedt2017,Jhuang:ICCV:2013}.
By addressing the interplay between data augmentation and skeleton-based gesture recognition, this work bridges a critical gap in the field, paving the way for scalable, robust, and real-world-ready gesture recognition solutions.
\section{RELATED WORK}
\subsection{Gesture Recognition}
Approaches for gesture recognition have traditionally relied on RGB images or depth data to capture movements, with convolutional neural networks (CNNs) extracting spatio-temporal features. While 2D CNNs were used for static gestures, 3D CNNs incorporated temporal dimensions for dynamic gestures \cite{simonyan2014twostream, ji20133dcnn}. However, image-based techniques face challenges such as lighting variations, occlusions, and cluttered backgrounds. To address these, multi-modal methods integrating depth, skeletal data, and point clouds have been developed, offering improved robustness by combining spatial and depth information \cite{wan2016handnet, oikonomidis2011full, neverova2015hand}.

e2eET Skeleton Based HGR Using Data-Level Fusion \cite{10216066} employed the learning from different data streams to represent different topologies of the same gesture, such as top-down, side-left, or front-away. By fine-tuning an ensemble of CNN streams in an end-to-end manner, the system achieves a deeper understanding of semantic representation and integration of the data results in more precise and robust hand gesture detections and recognition. The e2eET model is currently the state-of-the-art (SOTA) model on both the SHREC'17 \cite{DeSmedt2017} and DHG datasets \cite{DeSmedt2016}.
Point cloud gesture recognition is proposed by \cite{bigalke2021fusing} based on a dual stream framework which integrates raw point clouds and residual basis point sets (BPS) to address multi-scale challenges in gesture recognition. By fusing local and global spatial hand motion representations through a DenseNet and point network combination, their framework outperforms conventional methods on datasets such as DHG14/28 and SHREC’17 \cite{DeSmedt2016, DeSmedt2017}. The FPPR-PCD model ranks second in terms of performance on the SHREC'17 dataset.
\cite{min_CVPR2020_PointLSTM} Approached gesture recognition as an irregular sequence learning task and proposed a PointLSTM to model dynamic motion in point cloud sequences and propagate information from previous frames using neighbourhood grouping effectively to maintain spatial coherence. It was evaluated on SHREC’17 and NVGesture \cite{molchanov2016online} datasets, and showcast its ability to leverage both motion and shape features effectively. By overcoming the drawbacks of conventional Graph Convolutional Networks (GCNs), the Temporal Decoupling Graph Convolutional Network \cite{10113233} introduces temporal-dependent adjacency matrices, allowing temporal-sensitive topology learning. This approach, combined with channel-dependent adjacency matrices, provides an understanding of spatio-temporal relationships in gesture sequences.

DD-Net, a lightweight and efficient architecture for skeleton-based action recognition \cite{yang2019ddnet}. DD-Net addresses the issues of large model size and slow execution speed observed in prior approaches. By the formulation of new strategies like the Double-feature Double-motion strategy, where joint collection distances (JCD) for location-respective viewpoints were also combined with two scales of global motion features that portrayed fast and slow time variances. DD-Net was evaluated using SHREC’17 \cite{DeSmedt2017} and JHMDB \cite{Jhuang:ICCV:2013} datasets.
Advances in hardware, such as edge computing \cite{li2020edge}, and emerging paradigms like adversarial training \cite{goodfellow2014explaining}, have further enabled real-time gesture recognition in applications such as gaming, augmented reality, and medical assistive technologies. However, challenges remain, including occlusions and inter-user variability, which are being addressed through techniques like meta-learning \cite{finn2017model} and domain adaptation \cite{ganin2015unsupervised}. 
\subsection{Data Augmentation}
To address complex neural networks generalisation ability \cite{ABrain.HPGPT,ABrain.NNGPT,ABrain.NN-Dataset,Rupani2025llm,Gado2025llm}, different data augmentation techniques have been researched. SimplePairing \cite{inoue2018dataaugmentationpairingsamples} proposed an augmentation framework that investigates the impact of mixing two randomly selected images from the training set by averaging their pixel values, which gives N 2 new training samples, and showed a decrease in the error rate of the top-1 by 1. 29\% in CIFAR-10 \cite{krizhevsky2009cifar} and 4. 5\% in the ILSVRC 2012 dataset with GoogLeNet \cite{7298594}.
\cite{8795523} introduced a non-conventional augmentation technique by applying image cropping of four different images and patching them in one image to create a new training image and applying soft labelling to the cropped images classes. This approach achieved a test error of 2.19\% on CIFAR-10 \cite{krizhevsky2009cifar}.
To enhance the generalisation of neural networks, \cite{yun2019cutmix} proposed an augmentation technique, cutMix, which replaces regions of an image with patches from another while proportionally mixing their labels to retain informative pixels lost in traditional regional dropouts. The CutMix showed an improvement in top-1 accuracy on ImageNet classification of 2.28\% and 1.70\% on ResNet-50 and ResNet-101 \cite{7298594} architectures, respectively.

Data augmentation aims to imitate real-world scenarios and improve the generalisation ability of CNNs. The proposed augmentation in \cite{zhong2017randomerasingdataaugmentation} involves simulating occlusions commonly found in real-world environments by erasing a rectangular part of an image with random pixel values. This approach demonstrated effectiveness on the WRN-28-10 network architecture \cite{zagoruyko2017wideresidualnetworks} by 0.36\% in the top-1 error in Fashion-MNIST \cite{xiao2017fashionmnistnovelimagedataset} and an improvement in precision by 0.49\% using the ResNet-110 network architecture \cite{he2016identitymappingsdeepresidual}.

Complex augmentation techniques have been created recently, utilising automation and subject expertise to produce diverse and realistic data. Mixup \cite{zhang2018mixup} improves the generalisation of the model by producing augmented samples by linear interpolation between image pairings and their labels. Moreover, CutOut \cite{devries2017cutout} proposed a framework that enables the model to focus on global context instead of localised features by masking random rectangular portions of images.

\section{ IMPLEMENTATION}
\begin{figure}[ht!]
\centering
\resizebox{\columnwidth}{!}{%
\begin{tikzpicture}[node distance=2.5cm]

\node (orig_data) [dataset] {Original Datasets\ (SHREC'17, JHMDB, DHG)};
\node (aug_data) [dataset, right of=orig_data, xshift=8cm] {Generated Datasets};

\node (crop) [augment, below of=orig_data, xshift=2.5cm] {Random Crops};
\node (zoom) [augment, below of=crop] {Zooming};
\node (rotation) [augment, below of=zoom] {Rotations};
\node (brightness_contrast) [augment, below of=rotation] {Random Brightness and Contrast Adjustments};

\node [draw, thick, dashed, rounded corners, fit={(crop) (brightness_contrast)}, inner sep=0.75cm, label={[align=center]above:Augmentation Pipeline}] {};

\node (dd_net) [model, below of=brightness_contrast, xshift=-4.5cm] {DD-Net\ (Skeletal Input)};
\node (fppr) [model, below of=brightness_contrast, xshift=0cm] {FPPR\ (Point Cloud Input)};
\node (e2eet) [model, below of=brightness_contrast, xshift=4.5cm] {e2eET\ (Skeletal Input)};

\node (evaluation) [evaluation, below of=fppr, yshift=0cm] {Evaluation Metrics\ (Accuracy)};

\draw [arrow, bend left=20, line width=2pt, thick, red] (orig_data.south) to (crop.west);
\draw [arrow, line width=2pt, thick] (crop.south) -- (zoom.north);
\draw [arrow, line width=2pt, thick] (zoom.south) -- (rotation.north);
\draw [arrow, line width=2pt, thick] (rotation.south) -- (brightness_contrast.north);
\draw [arrow, bend right=20, line width=2pt, thick, blue] (brightness_contrast.east) to (aug_data.west);

\draw [arrow, line width=2pt, dashed] (aug_data.south) |- (dd_net.north);
\draw [arrow, line width=2pt, dotted] (aug_data.south) -- (fppr.north);
\draw [arrow, line width=2pt, dashed] (aug_data.south) |- (e2eet.north);

\draw [arrow, line width=2pt, thick, green!70!black] (dd_net.south) |- (evaluation.north west);
\draw [arrow, line width=2pt, thick, orange] (fppr.south) -- (evaluation.north);
\draw [arrow, line width=2pt, thick, purple] (e2eet.south) |- (evaluation.north east);

\node [below of=orig_data, yshift=-0.5cm] { Data Processed};
\node [below of=aug_data, yshift=-0.6cm] {Generated Data for Training};

\end{tikzpicture}%
}
\caption{Framework overview: The proposed augmentation framework improves training data and model performance across datasets and architectures.}
\label{fig:framework}
\end{figure}
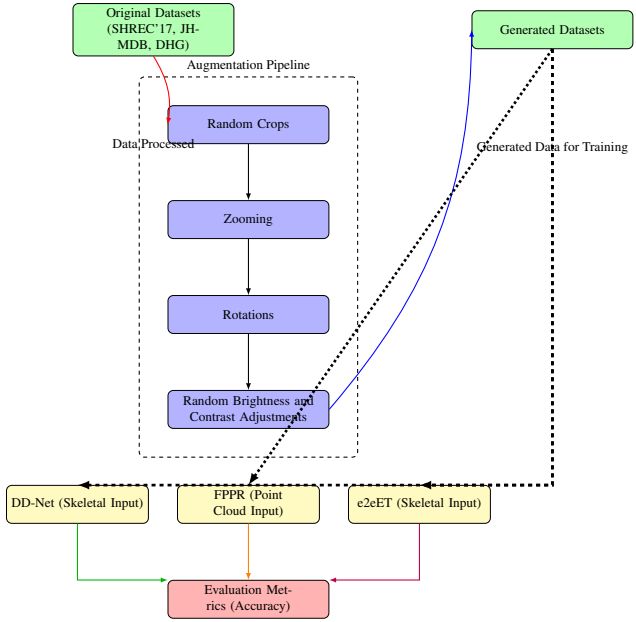

The proposed framework enhances gesture recognition by implementing a data augmentation pipeline that addresses challenges like limited dataset size, viewpoint bias, and environmental variations while preserving gesture-specific features. Figure~\ref{fig:framework} pipeline of the AugmentGest framework.

\subsection{AugmentGest Framework}
Let $x \in \mathbb{R}^{W \times H \times C}$ and $y$ denote a training image and its label, respectively. The goal of the proposed augmentation algorithm is to generate a new training sample $(\tilde{x}, \tilde{y})$ by applying a combination of transformations to a single input sample $(x, y)$. The generated training sample $(\tilde{x}, \tilde{y})$ is used to train the model with its original loss function. 

\begin{figure}[ht!]
    \centering
        \includegraphics[width=0.47\textwidth]{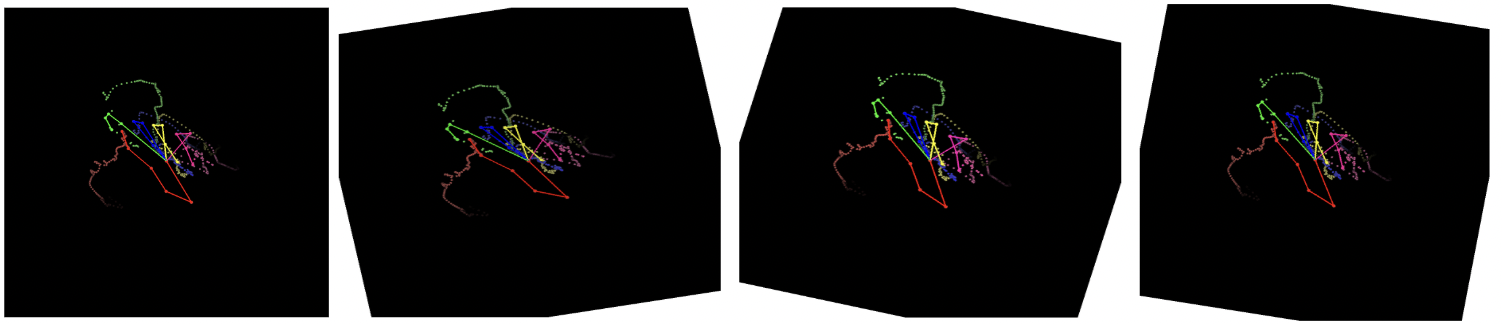}
        \caption{Results of Image Augmentation with the Proposed Framework.}
        \label{fig:shrecaug}
\end{figure}

The algorithm is designed to increase the training sample size by a factor of four, significantly enhancing the diversity of the dataset. The sequence of augmentation applied to the data is described as follows.
\begin{enumerate}
    \item \textbf{Random Cropping:} To preserve essential gesture information, the image is cropped to a size of $0.9W \times 0.9H$ or $0.95W \times 0.95H$, chosen randomly.
    \item \textbf{Random Rotation:} The cropped image rotates at an angle $\theta$ uniformly sampled from the range $[-15^\circ, 15^\circ]$. 
    \item \textbf{Zooming:} The rotated image is scaled by a zoom factor $\zeta$, sampled from a uniform range $[0.90, 1.10]$.
    \item \textbf{Brightness and Contrast Adjustment}: Finally, the brightness and contrast of the augmented image are adjusted by factors $\beta$ and $\gamma$, sampled from the ranges $[0.8, 1.2]$ and $[0.8, 1.2]$, respectively.
\end{enumerate}
The sequence of transformations can be represented as:
\[
\tilde{x} = T_{\text{brightness, contrast}}(T_{\text{zoom}}(T_{\text{rotation}}(T_{\text{crop}}(x))))
\]
where $T_{\text{crop}}$, $T_{\text{rotation}}$, $T_{\text{zoom}}$, and $T_{\text{brightness, contrast}}$ represent the cropping, rotation, zooming, and brightness/contrast transformations, respectively.
Figure \ref{fig:shrecaug} shows the result of applying the proposed framework to a "Grab" gesture sample from the SHREC’17 dataset. Using data-level fusion, depth images and 22 joint coordinates are combined into a static 2D spatiotemporal image representing hand joint positions. Temporal information is encoded through color intensity, as described in \cite{10216066}.
\subsection{Motivation and Comparison}
Unlike augmentation methods such as CutMix~\cite{yun2019cutmix}, MixUp~\cite{zhang2018mixup}, and SimplePairing~\cite{inoue2018dataaugmentationpairingsamples}, which mix training samples, our framework preserves gesture-specific spatial and temporal integrity. Random cropping ensures key regions are retained, while rotation, zooming, and brightness/contrast adjustments enhance robustness to variations in orientation, scale, and lighting. By maintaining structural coherence and quadrupling the training dataset, our approach provides a richer sample set for robust model training.
Table~\ref{tab:mix_results} compares MixUp and CutMix on the DD-Net model. While the baseline achieved 81.82\% accuracy, MixUp and CutMix resulted in 71.50\% and 72.50\%, respectively, showing performance drops of 10.80\% and 9.09\%. These methods, though effective in other contexts, disrupt the spatio-temporal coherence critical for gesture recognition.
\begin{table}[ht!]
\centering
\caption{Comparison of Data Augmentation Techniques on DD-Net Performance}
\label{tab:mix_results}
\begin{tabular}{|c|c|}
\hline
\textbf{Augmentation Technique} & \textbf{Accuracy (\%)}  \\ \hline
No Augmentation (Baseline)      & \textbf{81.82}          \\
MixUp                           & 71.02           \\ 
CutMix                          & 72.73           \\ 
\hline
\end{tabular}
\end{table}
\section{EVALUATION AND DISCUSSION}
\subsection{Datasets}
AugmentGest is evaluated on three benchmark datasets for hand gesture and human action recognition. SHREC’17 \cite{DeSmedt2017} contains 2800 gesture sequences across 14 and 28 gesture classes, while DHG14/28 \cite{DeSmedt2016} includes 2800 sequences across 14 gesture classes performed by 20 subjects. The JHMDB dataset \cite{Jhuang:ICCV:2013} comprises 928 action videos spanning 21 human action classes with annotated joint data. Together, these datasets provide a diverse environment for validating the proposed augmentation strategy.
\subsection{Experimental Setup}
The experiments were conducted on Kaggle Notebooks using cloud-based resources. \textbf{DD-Net} and \textbf{e2eET} were run without GPU acceleration, demonstrating their suitability for low-power environments, while \textbf{FPPR-PCD} required one GPU due to its computational complexity. A setup of two NVIDIA Tesla T4 GPUs (T4 x2) with CUDA acceleration was used, showcasing scalability and efficiency across different computational settings.
Table ~\ref{tab:results} compares the reproduced baseline performance with the results of the proposed augmentation framework. The published SOTA results are slightly higher, however, the reproduced results align with original implementations, providing a reliable baseline to assess the framework's impact.
\subsection{Evaluation on Gesture Recognition Frameworks}
The proposed framework integrates augmented datasets into various hand and motion gesture recognition models to assign a dynamic gesture sequence \( g_i \) to its correct class \( C_h \) within a set of gesture classes \( S = \{C_h\}_{h=1}^{N} \). The models evaluated include:
\begin{itemize}
    \item \textbf{E2eET} \cite{10216066}: A multi-stream CNN-based architecture using data-level fusion of gesture representations from multiple viewpoints. The evaluation reproduces the PyTorch implementation with the Adam optimizer and homoscedastic cross-entropy loss, leveraging ResNet34. Accuracy was monitored for top-down, custom, and front-away viewpoints.
    \item \textbf{DD-Net} \cite{yang2019ddnet}: A lightweight model optimized for learning spatial and temporal features from 3D skeletal joint data, trained with categorical cross-entropy loss and the Adam optimizer.
    \item \textbf{FPPR-PCD} \cite{bigalke2021fusing}: A dual-stream point cloud-based model that decouples local posture and global spatial features. Depth maps were converted into point cloud sequences using the \texttt{dbscanCluster} format.
\end{itemize}
This setup reflects a carefully optimised pipeline tailored to enhance performance across diverse architectures, validating its versatility and effectiveness in improving gesture recognition accuracy.
\subsection{Results and Discussion}
\begin{figure}[ht!]
    \centering
        \includegraphics[width=0.44\textwidth]{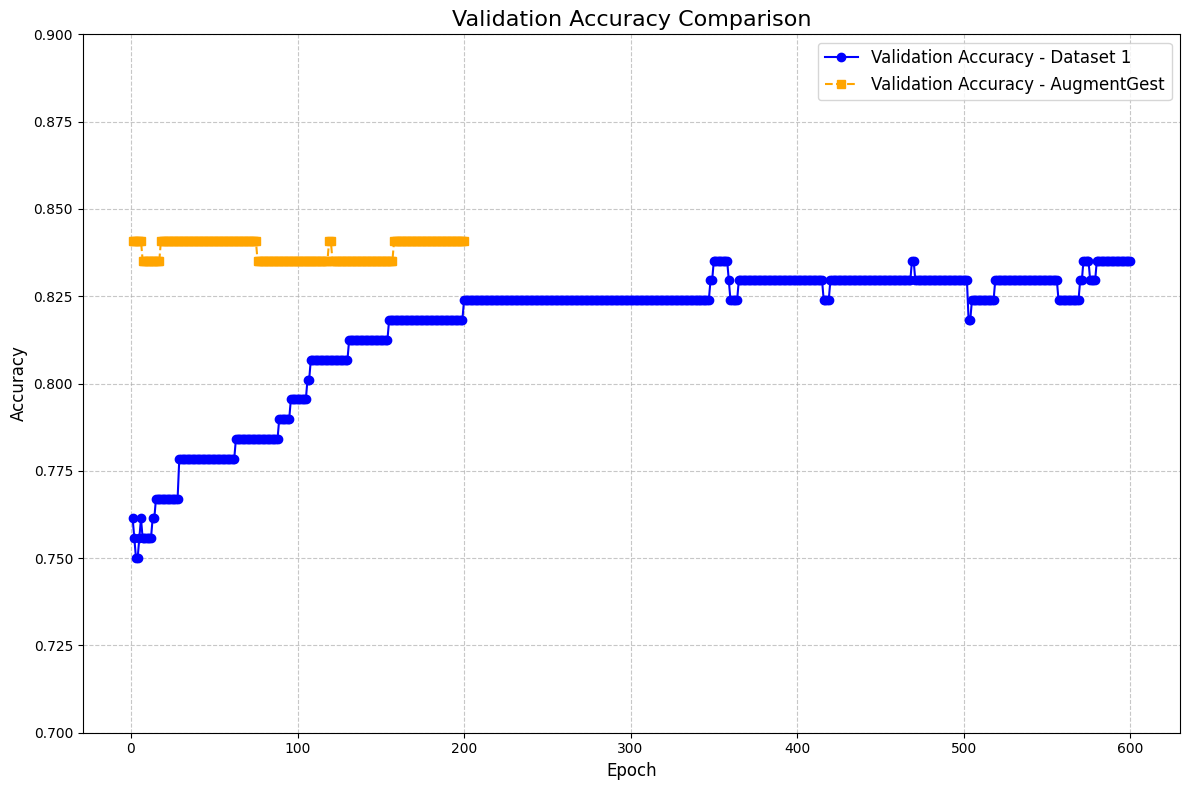}
        \caption{DD-Net Validation accuracy trained on different datasets: JHMDB dataset (Dataset 1) and an augmented dataset (AugmentGest).}
        \label{fig:GestAugment_vs_orig}
\end{figure}

The proposed framework expanded dataset size and diversity by introducing variations in scale, orientation while preserving key gesture features.
\begin{table}[ht!]
\centering
\caption{Comparison of reproduced model performance on different datasets, with and without the proposed augmentation framework.}
\label{tab:results}
\begin{tabular}{|l|l|c|c|}
\hline
\textbf{Model}         & \textbf{Dataset}    & \textbf{Accuracy} & \textbf{Accuracy} 
\\
& & (\%)&  AugmentGest
\\
\hline
\multirow{2}{*}{DD-Net} & SHREC 14g          & 94.76                        & \textbf{95.48}                      \\ 
                        & JHMDB              & 81.82                        & \textbf{86.36}                      \\ 
\hline
FPPR-PCD                   & SHREC 14g          & 95.90                        & \textbf{96.40}                      \\ 
\hline
\multirow{3}{*}{e2eET}  & SHREC 14g          & 96.67                      & \textbf{98.21}                      \\ 
                        & SHREC 28g          & 94.05                        & \textbf{94.52}                      \\ 
                        & DHG 28g            & 91.67                        & \textbf{92.98}                      \\ 
\hline
\end{tabular}
\\
\vspace{0.7em}
\small
\raggedright
\textbf{Note:} SHREC 14g and 28g refer to SHREC'17 14-gesture and 28-gesture subsets of the dataset. The red values represent the improved accuracy achieved using the proposed augmentation framework.
\end{table}
For the SHREC'17 14G dataset, the state-of-the-art e2eET model demonstrates significant performance improvements when trained with the augmentation data sample with AugmentGest. As shown in table \ref{tab:results} the e2eET model achieves a 1.54\% accuracy increase, surpassing the previously reported SOTA results in \cite{10216066}. 
The framework delivers improvements of 0.47\% and 1.31\% on the SHREC'17 28-gesture and DHG 28-gesture datasets, respectively. 
Furthermore, the FPPR model, benefits from the AugmentGest framework, achieving an increase in accuracy on the SHREC'17 14-gesture dataset by 0.5\%. This result underscores the framework's adaptability to diverse gesture recognition framework architectures.
The results in Table \ref{tab:results} demonstrate consistent performance improvements across all models and datasets. Notably, the DD-Net model achieved a 4.54\% accuracy increase on the JHMDB motion dataset and improved from 94.76\% to \textbf{95.48\%} on the SHREC'17 14-hand gesture dataset. These findings validate the effectiveness of the proposed augmentation framework in enhancing gesture recognition performance across diverse datasets and architectures, highlighting its versatility and potential for various applications.
Figure \ref{fig:GestAugment_vs_orig} illustrates the validation accuracy of the model trained on the SHREC'17 14G dataset versus the augmented dataset. The dataset generated through AugmentGest framework is four times larger than the original dataset, allowing for a more robust evaluation of the model's performance. For evaluation, the model was trained for 200 epochs on the augmented dataset, as opposed to 600 epochs on the original dataset, reflecting the reduced need for extended training due to the increased diversity and volume of training samples. The results demonstrate that the augmented dataset achieves consistently higher validation accuracy within fewer epochs, underscoring the effectiveness of the augmentation framework in improving model generalization and efficiency.
Furthermore, extended training on the augmented dataset for 600 epochs improves accuracy to 86\%.
\subsection{Ablation Study}
\begin{table}[h!]
\centering
\caption{Ablation study of the Augmentation Pipeline to show the contribution of each step to the E2eET Multi-stream CNN model's performance on SHREC'17.}
\label{tab:ablation}
\begin{tabular}{|l|c|}
\hline
\textbf{Framework Step}        & \textbf{Accuracy (\%) } \\
\hline
No Augmentation (Baseline) & 96.67     \\ 
Image Crops                &  \textbf{97.50}   \\ 
Image Rotations            &  \textbf{97.50}    \\ 
Image Zoom                 &     96.55       \\ 
Brightness \& Contrast Adjustment                &     \textbf{97.14}     \\ 
\hline
\end{tabular}
\end{table}
To further investigate the impact of the proposed framework, individual augmentation steps on the E2eET model \cite{10216066} were evaluated using the SHREC'17 dataset, as shown in Table \ref{tab:ablation}. The cropping and rotation of the images improved each by 0.83\%, and brightness \& contrast adjustment improved by 0.47\%, while the image zoom had no effect. Combining all steps yielded an accuracy of 98.20\%, highlighting the complementary benefits of augmentation techniques.
\subsection{ Evaluation of Training Time}
\begin{table}[ht!]
\centering
\caption{Runtime of the E2eET Model.}
\label{tab:training}
\begin{tabular}{|l|c|c|}
\hline
\textbf{Dataset} & \textbf{Original } & \textbf{Augmented } \\

\hline
SHREC'17 14g     & 03h:04m:26s                   & 03h:10m:42s                     \\ 
SHREC'17 28g     & 02h:53m:45s                   & 03h:09m:07s                     \\ 
DHG 28g          & 02h:54m:37s                   & 02h:52m:53s                     \\ \hline
\end{tabular}
\end{table}
Table \ref{tab:training} presents the training and evaluation runtime of the 2eET Skeleton Based HGR on the datasets compared to the datasets generated using the AugmentGest framework. Despite the augmented datasets being four times larger, the increase in runtime is minimal. 
For the SHREC'17 datasets 14G and 28G, the runtime increases by only a few minutes, while for the DHG 28-gesture dataset, the runtime remains nearly unchanged. These results demonstrate the computational efficiency of the proposed augmentation framework, making it a practical solution for significantly expanding dataset size without incurring substantial overhead.

\section{CONCLUSION \& FUTURE WORK}
This paper introduced a novel augmentation framework designed to enhance the performance of gesture recognition models. By preserving spatiotemporal coherence and skeleton features, the framework improves accuracy without additional data collection costs. Integrating diverse augmentation strategies, it demonstrated significant accuracy gains across state-of-the-art models such as e2eET, DD-Net, and FPPR-PCD on benchmark datasets like SHREC’17 and DHG. These results underscore the framework's ability to enhance model generalization and efficiency, even with datasets of varying complexities and gesture diversity.
While the framework showcases its potential, challenges remain, particularly in addressing real-world variability and expanding its application to more diverse datasets. Future research could focus on context-aware augmentation techniques and attention-driven methods to further refine and optimise the framework. By tackling these challenges, this approach can pave the way for more robust and adaptable gesture recognition systems suitable for dynamic and complex environments.
The findings of this study underscore the critical role of data augmentation in advancing gesture recognition, contributing to improved robustness and accuracy across a range of applications, including gaming, virtual reality, and assistive technologies.
{\small
\bibliographystyle{ieee}
\bibliography{main}
}

\end{document}